\documentclass[letterpaper]{article} 
\usepackage{aaai2026}  
\usepackage{times}  
\usepackage{helvet}  
\usepackage{courier}  
\usepackage[hyphens]{url}  
\usepackage{graphicx} 
\usepackage{natbib}  
\usepackage{caption} 
\usepackage{mdframed}
\usepackage{algorithm}
\usepackage{lipsum}
\usepackage{algpseudocode}
\usepackage{amsmath}
\usepackage{amssymb}
\usepackage{booktabs}
\usepackage[table]{xcolor} 
\usepackage{mdframed}
\usepackage{tcolorbox}

\definecolor{backcolor}{gray}{0.85}

\usepackage{newfloat}
\usepackage{listings}
\DeclareCaptionStyle{ruled}{labelfont=normalfont,labelsep=colon,strut=off} 
\floatstyle{ruled}
\newfloat{listing}{tb}{lst}{}
\floatname{listing}{Listing}
\title{A Principle-Driven Adaptive Policy for Group Cognitive Stimulation \\ Dialogue for Elderly with Cognitive Impairment}
\author{
    Jiyue Jiang\textsuperscript{\rm 1,}\thanks{These authors contributed equally to this work.},
    Yanyu Chen\textsuperscript{\rm 1,$\ast$}, Pengan Chen\textsuperscript{\rm 1}, Kai Liu\textsuperscript{\rm 2}, Jingqi Zhou\textsuperscript{\rm 2}, Zheyong Zhu\textsuperscript{\rm 1}, He Hu\textsuperscript{\rm 3},\\
    Fei Ma\textsuperscript{\rm 3,}\thanks{Corresponding authors: mafei@gml.ac.cn, cwu@cs.hku.hk.}, Qi Tian\textsuperscript{\rm 3,4}, Chuan Wu\textsuperscript{\rm 2,$\dagger$}
}
\affiliations{
    \textsuperscript{\rm 1}The Chinese University of Hong Kong\\
    \textsuperscript{\rm 2}The University of Hong Kong\\
    \textsuperscript{\rm 3}Guangdong Laboratory of Artificial Intelligence and Digital Economy (SZ)\\
    \textsuperscript{\rm 4}Huawei\\
    \{jiangjy, chenyanyu.cse\}@link.cuhk.edu.hk, mafei@gml.ac.cn, cwu@cs.hku.hk
}

\usepackage{bibentry}

\begin{document}

\maketitle

\begin{abstract}
Cognitive impairment is becoming a major public health challenge. Cognitive Stimulation Therapy (CST) is an effective intervention for cognitive impairment, but traditional methods are difficult to scale, and existing digital systems struggle with group dialogues and cognitive stimulation principles. While Large Language Models (LLMs) are powerful, their application in this context faces key challenges: cognitive stimulation dialogue paradigms, a lack of therapeutic reasoning, and static-only user modeling. To address these issues, we propose a principle-driven adaptive policy actualized through a Group Cognitive Stimulation Dialogue (GCSD) system. We first construct a dataset with over 500 hours of real-world CST conversations and 10,000+ simulated dialogues generated via our Principle-Guided Scenario Simulation strategy. Our GCSD system then integrates four core modules to overcome LLM limitations: (i) a multi-speaker context controller to resolve role confusion; (ii) dynamic participant cognitive state modeling for personalized interaction; (iii) a cognitive stimulation-focused attention loss to instill cognitive stimulation reasoning; and (iv) a multi-dimensional reward strategy to enhance response value. Experimental results demonstrate that GCSD significantly outperforms baseline models across various evaluation metrics. Future work will focus on long-term clinical validation to bridge the gap between computational performance and clinical efficacy.
\end{abstract}


\section{Introduction}

\begin{figure}[h!]
    \centering
    \includegraphics[width=\linewidth]{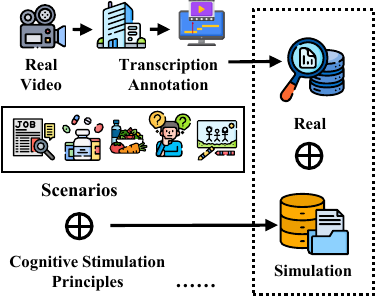}
    \caption{The construction of real and simulated datasets.} 
    \label{fig:1}
\end{figure}

As the global population ages, cognitive impairment has become a significant public health challenge, affecting over 55 million people worldwide~\cite{chib2025interventions}. Cognitive stimulation therapy~\cite{morley2014cognitive}, particularly in group settings~\cite{sun2022comparative, zhang2025constructive}, is an effective non-pharmacological intervention that maintains or improves cognitive function by fostering social interaction and collective reminiscence~\cite{spector2003efficacy}. Despite the proven efficacy of CST, the previous models are constrained by a reliance on professional therapists, fixed schedules, and physical locations, which makes them costly and difficult to scale.

Previous cognitive training systems have largely been confined to rigid, question-and-answer formats, making it difficult for them to generate natural, open-ended dialogue~\cite{magyar2019autonomous, hirose2020study, tokunaga2021dialogue, kim2021efficacy}. Due to their limited model scales, they also lack the capacity to learn and internalize the complex, multi-dimensional therapeutic principles behind the 18 cognitive stimulation therapy~\cite{woods2006improved, jiang2023cognitivestimulationdialoguemultisource}. Critically, these systems are entirely incapable of handling complex interactive scenarios such as multi-party dialogue, fundamentally neglecting the core therapeutic elements of group therapy, such as social interaction and peer support.

While large language models have brought revolutionary breakthroughs in recent years~\cite{jiang2025biological, jiang2025benchmarking, wang2025large, jiangetal2025well}, they still face three core challenges when applied to principle-driven multi-party/group cognitive stimulation dialogue scenarios:
(1) Mismatched Dialogue Paradigm: The core architectures and training paradigms of mainstream LLMs are predominantly designed for dyadic (two-person) interactions, leading to issues such as speaker confusion and context loss when handling multi-party interactions.
(2) Lack of Cognitive Stimulation Reasoning: As general-purpose models, they can generate fluent and empathetic text, but they lack the deep, cognitive stimulation-oriented reasoning and strategic guidance required by CST.
(3) Static User State Modeling: Existing LLMs struggle to dynamically and continuously model the cognitive state of each elder, a capability that is crucial for providing personalized and adaptive cognitive stimulation.

This paper introduces a principle-driven adaptive policy designed to guide a dialogue system that simulates the cognitive stimulation process of group cognitive stimulation, and enables ``anytime, anywhere'' access to cognitive stimulation through its digital implementation. To build such a system, we first construct a Cantonese multi-party dialogue dataset by meticulously transcribing and reviewing about 500 hours of real group cognitive stimulation video recordings in collaboration with a third-party company (responsible for the data transcription and annotation). Subsequently, we design the Principle-Guided Scenario Simulation (PGSS) strategy to generate extensive data. This dual-data method serves two purposes: (1) It enriches the diversity of training scenarios by covering specific themes and interaction patterns that are sparse or absent in the real data. (2) It allows the model to pre-learn foundational patterns, enabling it to better grasp the basic framework and principles of multi-party cognitive stimulation dialogue.

Our key contributions are as follows: 
\begin{itemize}
    \item We construct a large-scale dataset with over 500 hours of real-world Cantonese conversations and 10,000+ principle-guided simulated dialogues, providing a key resource for multi-party therapeutic dialogue research.
    \item We introduce GCSD, a principle-driven adaptive policy for multi-party cognitive stimulation dialogues. It integrates four core modules: a multi-speaker context controller, dynamic participant state modeling, a cognitive stimulation-focused attention loss, and a multi-dimensional reward algorithm to align responses with CST principles.
    \item Extensive evaluations demonstrate that GCSD significantly outperforms strong commercial and open-source baselines across automatic and human metrics, confirming the effectiveness of our proposed framework.
\end{itemize}

\section{Related Work}
\subsection{Cognitive Training Dialogue Systems}
Recent advancements in dialogue systems have shifted from task-oriented designs to open-domain chatbots powered by LLMs \cite{lee2016quote, su14031520, ni2023recent, sanchez2024automating}. While significant progress has been made in English-language systems~\cite{yi2024survey, pan2025ellma}, Chinese dialogue systems have also seen notable developments in general-purpose applications~\cite{zhou2021eva, gu2023eva2}. However, existing cognitive training dialogue systems often overlook the unique needs of elderly users with cognitive impairment. Current solutions, including robotic photo-based interactions and metamemory therapy~\cite{magyar2019autonomous, hirose2020study, tokunaga2021dialogue, kim2021efficacy}, provide limited cognitive stimulation. Although recent work has introduced specialized datasets for Chinese cognitive training~\cite{jiang2023cognitivestimulationdialoguemultisource}, their effectiveness remains constrained, particularly for low-resource languages like Cantonese. 



\section{Data Construction}

\subsection{Real Data Construction}

Cognitive stimulation therapy is an evidence-based, non-pharmacological intervention for elders with mild to moderate cognitive impairment~\cite{morley2014cognitive, zhang2025constructive}. It utilizes themed group sessions to enhance cognitive function, memory, and quality of life.

We process a dataset of about 500 hours of Cantonese group cognitive stimulation video data. The data undergoes manual transcription and annotation by a third-party company. To clean the spoken text, we apply rule-based processing: (1) truncating text exceeding 1000 characters; (2) removing special characters like [S] and [UNK]; (3) retaining only the final symbol in a series of punctuation marks.

To construct multi-party dialogue data, we insert an [Assistant] token before each therapist's utterance and a [Human\_i] token (where ``i'' represents a unique individual) before each elder's utterance. This process yields structured, authentic multi-party cognitive stimulation dialogue data.

\begin{table}[!t]\small
\centering
\begin{tabular}{l|c|c}
\toprule
\textbf{Categories} & \textbf{Num. (Real)} & \textbf{Num. (Stimulation)}\\
\midrule
\midrule
    \rowcolor{backcolor}
    \multicolumn{3}{c}{\textsc{Training Dataset}} \\
\midrule
\textbf{Data Count} & 10,502&9,429\\
\textbf{Total Tokens} &407,699& 295,902\\
Assistant Count &160,619& 119,370\\
Assistant Tokens &1,644& 119,511\\
Human/Elder Number &9& 7\\
Total Avg Length &38.82& 31.38\\
Assistant Avg Length &1.024& 1.001 \\
Human Avg Length &1.006& 1.001\\
\midrule
\midrule
    \rowcolor{backcolor}
    \multicolumn{3}{c}{\textsc{Testing Dataset}} \\
\midrule
\textbf{Data Count} &1,166& 1,047\\
\textbf{Total Tokens} &45,848& 32,862\\
Assistant Count &18,083& 13,238\\
Assistant Tokens &18,515& 13,262\\
Human/Elder Number &9& 6\\
Total Avg Length &39.32& 31.39\\
Assistant Avg Length &1.024& 1.002 \\
Human Avg Length &1.006& 1.000\\
\bottomrule
\end{tabular}
\caption{Statistics of the multi-party cognitive stimulation dialogue dataset.}
\label{TableB}
\end{table}

\subsection{Simulated Data Construction - Principle-Guided Scenario Simulation Strategy}

To address the sparsity of specific themes and interaction patterns in real-world data, and to enable the model to pre-learn the foundational frameworks and cognitive stimulation tenets of multi-party dialogue, we construct a large-scale simulated dataset. For this purpose, we design and implement the principle-guided scenario simulation strategy to generate high-quality dialogues using the GPT-4o API\footnote{\url{https://openai.com/index/hello-gpt-4o/}}.

The core of the PGSS strategy is the construction of a detailed prompt, a process that follows these steps (the complete prompt is provided in \textit{Appendix Prompt}):

\begin{itemize}
    \item \textbf{Define Task and Output Format:} We first define the model's generation task: to produce multi-party dialogues exceeding 30 turns. We specify that the output must be a structured JSON object containing three fields: turn, speaker, and dialogue, ensuring data consistency and usability.
    \item \textbf{Establish Roles and Scenarios:} We establish the participant roles, comprising one therapist responsible for facilitation and five to six patients with cognitive impairments, each assigned a unique personal history and interests. The scenario is set in a well-equipped activity room to support diverse interactions.
    \item \textbf{Provide Activities and Prompts:} We provide a list of common activity categories from CST (art creation, thematic discussion, etc.) and instruct the model to randomly select from this list. In addition, we supply dialogue prompts for the opening, middle, and closing phases of a session to guide the model in generating conversations that align with the cognitive stimulation flow.
    \item \textbf{Explicitly List Constraining Principles:} The most critical step is that we explicitly list the 18 CST principles within the prompt (``Encourage new ideas'', ``Value opinions'', ``Use reminiscence''). We then require that the model's generation, particularly the therapist's facilitative utterances, must strictly adhere to these listed principles throughout the entire process.
\end{itemize}

The statistics of the real data and the simulated data (training and testing dataset) are presented in Table~\ref{TableB}.

\section{Methodology: GCSD}

\subsection{Overview}

\begin{figure*}[h!]
    \centering
    \includegraphics[width=\linewidth]{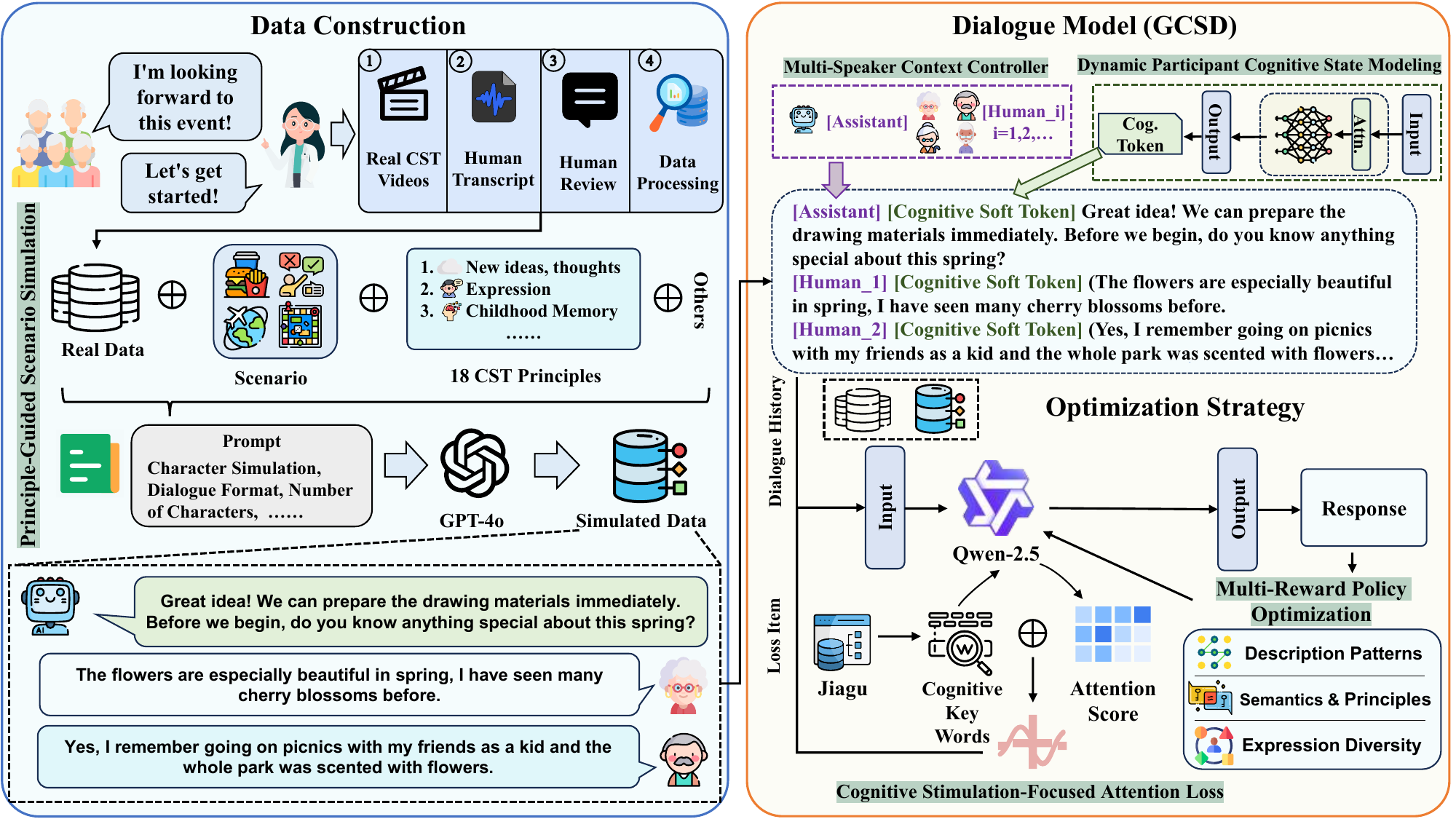}
    \caption{This is an overview of the GCSD. On the left side is the data construction phase, which includes the workflows for both real data and simulated data construction (principle-guided scenario simulation strategy), as well as an example shown at the lower left. On the right side is the GCSD model phase, where the model first learns patterns and scenarios based on simulated data, and then trains on real data. The model consists of four parts: multi-speaker context controller, dynamic participant cognitive state modeling, cognitive stimulation-focused attention Loss, and multi-reward policy optimization.} 
    \label{overview}
\end{figure*}

Figure~\ref{overview} gives an overview of our principle-driven adaptive policy, which is instantiated by the GCSD system. This policy is operationalized through four key stages: (1) multi-party context control; (2) dynamic participant cognitive state modeling; (3) cognitive stimulation-focused attention loss; and (4) multi-reward policy optimization. 

\subsection{Multi-Party Context Control}

Effectively managing speaker roles is a foundational requirement for coherent multi-party dialogue. To enable our model to learn the distinct characteristics of multiple elders, we adopt an established data formatting approach by employing special tokens. Specifically, we use \texttt{[Assistant]} to delineate utterances from the therapist and \texttt{[Human\_i]} (\texttt{[Human\_1]}, \texttt{[Human\_2]}) for each elder. For instance, a typical dialogue segment is structured as follows (\texttt{[|im\_start|>} is the start symbol, and \texttt{<|im\_end|>} is the end symbol): 


\begin{tcolorbox}[
    colback=blue!5!white,  
    colframe=blue!60!black, 
    boxrule=1pt,           
    arc=8pt,               
    left=10pt, right=10pt, top=8pt, bottom=8pt 
]
    \texttt{\textbf{<|im\_start|>}} \texttt{\textbf{[Human\_1]}} Mid-Autumn Festival is coming, which reminds me of the mooncakes we used to eat.\\
    \texttt{\textbf{[Human\_2]}} Yes, and we used to play with lanterns when we were young. \texttt{\textbf{<|im\_end|>}}\\
    \texttt{\textbf{[Assistant]}} Both are wonderful memories. If you had to choose one, which do you feel better represents the festival? \texttt{\textbf{<|im\_end|>}}\\
    \texttt{\textbf{[Human\_2]}} I would choose the lanterns, because my whole family would make them together. \texttt{\textbf{<|im\_end|>}}  \textbf{......}
\end{tcolorbox}

This method ensures the model can accurately attribute utterances to speaker, maintaining role consistency and contextual coherence. In addition, we adopt a two-stage training strategy: first, continue training on the simulated dataset to learn structured, principle-aligned dialogue flows via PGSS; then, fine-tuning on real-world data to capture authentic linguistic nuances and conversational dynamics. 

\subsection{Dynamic Participant Cognitive State Modeling}

To overcome the static user state modeling limitations of standard LLMs, the dynamic participant state modeling module generates a dynamic soft prompt for each participant. This prompt, a continuous vector, is produced by a compact neural network designed to distill complex user state information (i.e., cognitive state, historical interactions) into a concise representation. This network operates in parallel to the dialogue model, dynamically modulating the generation to achieve personalized and adaptive responses.

\subsubsection{Soft Prompt Generation Network.}
The generator is an efficient multi-layer perceptron (MLP) that takes a concatenated input vector $X_{in}$ 
comprising various user state features. The network architecture is as follows:

\begin{itemize}
    \item \textbf{Input Layer:} A linear layer transforms the input features $X_{in}$ into a latent representation $L_1$, followed by a GELU activation~\cite{hendrycks2016gaussian}.
    \begin{equation}
    L_1 = \text{GELU}(X_{in} W_1 + b_1)
    \end{equation}
    where $W_1 \in \mathbb{R}^{D_{in} \times 512}$. $D_{in}$ is the dimension of the concatenated input features.

    \item \textbf{Hidden Layers:}\,  
The attention block and the final projection can be written compactly in two balanced lines:
\begin{equation}
\begin{aligned}
L_2 = \operatorname{GELU}\!\Bigl(&
      \operatorname{softmax}\!\Bigl(
        \frac{(L_1 W_Q)\,(L_1 W_K)^{\mathsf T}}{\sqrt{d_k}}
      \Bigr)\\[4pt]
      &
      (L_1 W_V)\,W_2 \;+\; b_2
\Bigr)
\end{aligned}
\end{equation}

where  
$Q = L_1 W_Q,\;K = L_1 W_K,\;V = L_1 W_V$ with $W_Q, W_K, W_V \in \mathbb{R}^{512 \times d_k}$ ($d_k = 512$),  
and the output layer parameters are $W_2 \in \mathbb{R}^{512 \times 256}$ and $b_2 \in \mathbb{R}^{256}$.
        
    
    \item \textbf{Output Layer:} A final linear layer produces the soft prompt vector $P_{\text{soft}} \in \mathbb{R}^{D_{\text{prompt}}}$, where the dimension $D_{\text{prompt}}$ matches the main model's embedding dimension (i.e., 512). A $\tanh$ activation function bounds the output values, enhancing training stability.
    \begin{equation}
        P_{\text{soft}} = \tanh(L_2 W_{\text{out}} + b_{\text{out}})
    \end{equation}
    
    \noindent
    where $W_{\text{out}} \in \mathbb{R}^{256 \times D_{\text{prompt}}}$.
\end{itemize}

\subsubsection{Integration and Smoothing.}
The generated soft prompt $P_{\text{soft}}$ is prepended to the input token embeddings of the main Transformer-based dialogue model at each decoding step:
\begin{equation}
    \text{Input}_{\text{Transformer}} = [P_{\text{soft}}; E_{\text{token}_1}; E_{\text{token}_2}; \dots; E_{\text{token}_k}]
\end{equation}
where $E_{\text{token}_i}$ are the learned embeddings of the input tokens. By treating $P_{\text{soft}}$ as a ``virtual token'' embedding, it influences the self-attention mechanisms, conditioning the model's output on the elder's dynamically updated state.

To prevent erratic fluctuations in the prompt, which is critical when interacting with a vulnerable population, we apply a temporal smoothness regularization term:
\begin{equation}
    \mathcal{L}_{\text{Smoothness}} = ||P_{\text{soft},t} - P_{\text{soft},t-1}||_2^2
\end{equation}
This ensures that the personalized guidance evolves smoothly over the course of the interaction.

\subsection{Optimization Strategy}

Our training process is divided into two core phases: (1) Supervised Fine-Tuning (SFT), which teaches the model to generate fluent, coherent, and therapeutically-focused responses; and (2) Policy Optimization, which further aligns the model's behavior with multi-dimensional cognitive stimulation objectives using a reward-based algorithm.

\subsubsection{Phase 1: Supervised Fine-Tuning with Cognitive Stimulation-Focused Attention.}

In the supervised fine-tuning phase, we optimize the model using a joint loss function composed of three components:

\noindent
\textbf{1. Generation Loss ($\mathcal{L}_{\text{Gen}}$):} This is the standard auto-regressive cross-entropy loss, ensuring linguistic fluency and contextual relevance~\cite{radford2019language, yang2025qwen2}.

\begin{equation}
\mathcal{L}_{\text{gen}} = -\sum_{i=1}^{N} \log P(y_i | y_{<i}, \mathcal{U})
\end{equation}

\noindent
where $y_i$ is the target token at step $i$, given the context $\mathcal{U}$ and the preceding tokens $y_{<i}$. $N$ is the total sequence length.

\noindent
\textbf{2. Cognitive Stimulation-Focused Attention Loss ($\mathcal{L}_{\text{CSFAL}}$):} To instill cognitive stimulation reasoning, we design a loss to guide the model's attention mechanism. We first use a keyword extraction tool\footnote{https://github.com/ownthink/Jiagu} to identify keywords from the reference response. We then compute a weighted mean squared error between the model's attention weights and a target distribution derived from these keywords.

\begin{equation}
\mathcal{L}_{\text{CSFAL}} = \frac{1}{M} \sum_{j=1}^{M} \lambda_j (a_j - \eta_j)^2
\end{equation}

\noindent
where $M$ is the length of the token sequence; $a_j$ is the model's attention weight on token $j$ (averaged across heads); $\eta_j$ is the target attention score for token $j$. If the token is an extracted keyword, $\eta_j > 0$; otherwise, it is 0; $\lambda_j$ is a saliency weight, such as $\exp(\kappa \cdot \eta_j)$, that amplifies the penalty for critical items, where $\kappa$ is a tunable hyperparameter.

\begin{table*}[h]
\label{0shot}
\small
\centering
\begin{tabular}{l|c|c|c|c|c|c|c|c}
\toprule
\textbf{Models} & \textbf{ROUGE-L} & \textbf{BLEU-2} & \textbf{BLEU-4} & \textbf{BERTScore} & \textbf{Distinct-2} & \textbf{Relevance} & \textbf{Empathy} & \textbf{Fluency}\\
\midrule
\midrule
    \rowcolor{backcolor}
    \multicolumn{9}{c}{\textsc{Simulated Dataset}} \\
\midrule
ERNIE   & 17.24 & 8.32& 7.38 & 65.35 &  53.92   &3.05&3.20&2.90    \\
Doubao-Pro   & 23.58 & 22.39&21.51 & 73.19&  71.03   &4.00&3.55&3.33    \\
GPT-4o   & 21.78 &20.38& 20.05 & 72.46&  67.83   &3.67&3.65&3.45    \\
DeepSeek-671B   & 21.67 & 22.13 & 22.09 & 77.29 &  \textbf{73.79}   &3.59&3.45&3.36    \\
LLaMA-3.1-405B   & 22.56 & 22.13& 21.08 & 76.52& 71.26  &3.75&3.70&3.42    \\
\midrule
GLM-4-Plus (5-Shot)   & 17.69 & 18.24 & 18.11 & 70.18&  72.97   &3.45&3.15&3.15    \\
ChatGPT (5-Shot)   & 21.57 & 18.06 & 15.44 & 72.18&  52.86   &3.95&3.50&3.33    \\
DeepSeek-671B (5-Shot)   & 22.89 & 24.22 & 23.59 & 76.26 &  72.31   &4.02&3.57&3.40    \\
Qwen-Max (5-Shot)   & 21.54 & 21.34& 20.15 & 74.27& 64.41  &3.65&3.60&3.38    \\
\midrule
GCSD w/o DPSM & 21.53 & 24.45 & 22.18 & 79.33 & 56.41 & 3.60 & 3.54 & 3.32 \\
GCSD w/o CSFAL & 23.58 & 25.16 & 22.64 & 74.19 & 68.42 & 3.58 & 3.58 & 3.33\\
\textbf{GCSD-3b}   & \textbf{26.92}&\textbf{28.32}& \textbf{26.83}&\textbf{79.45}&  72.64&\textbf{4.20}&\textbf{3.67}&\textbf{3.50}     \\
\midrule
\midrule
    \rowcolor{backcolor}
    \multicolumn{9}{c}{\textsc{Real Dataset}} \\
\midrule
ERNIE   & 18.32 & 11.32& 9.16 & 66.93 &  56.28   &3.33&2.95&3.00    \\
Doubao-Pro   & 24.87 & 23.59 & 22.36 & 75.16&  71.46   &3.95&3.33&3.20    \\
GPT-4o   & 25.76 & 21.53& 20.14 & 73.79&  69.15   & 4.00 & 3.45& 3.35    \\
DeepSeek-671B   & 22.54 & 23.58 & 22.42 & 79.98 &  \textbf{76.86}   &4.08&3.45&3.46    \\
LLaMA-3.1-405B   & 23.87 & 23.68& 23.52 & 77.13& 71.52  &4.00&3.45&3.40    \\
\midrule
GLM-4-Plus (5-Shot)   & 18.54 & 18.53 & 17.39 & 70.32&  72.18   &3.50&3.27&3.33    \\
ChatGPT (5-Shot)   & 22.32 & 18.23 & 16.19 & 73.54&  55.38   & 3.67& 3.33&3.25    \\
DeepSeek-671B (5-Shot)   & 23.58 & 24.83 & 24.36 & 78.27 &  73.27   &4.10&3.48&3.42    \\
Qwen-Max (5-Shot)  & 22.87 & 21.56& 20.53 & 73.63 & 64.76  &3.67&3.40&3.38    \\
\midrule
GCSD w/o CT & 24.33 & 27.14 & 26.51 & 75.42 & 74.19 &3.99 & 3.45 & 3.40\\
GCSD w/o DPSM & 23.57 & 25.63 & 23.15 & 79.48 & 60.13 & 3.74 & 3.40 & 3.32 \\
GCSD w/o CSFAL & 25.18 & 26.16 & 24.98 & 76.33 & 70.46 & 3.69 & 3.39 & 3.36 \\
\textbf{GCSD-3b}   & \textbf{27.63}&\textbf{28.53}&\textbf{27.93}&\textbf{80.12} &74.82  & \textbf{4.15}  & \textbf{3.50} &  \textbf{3.53}   \\
\bottomrule
\end{tabular}
\caption{Evaluation results between baselines and our GCSD on the \textbf{cognitive stimulation simulated and real testing datset}. The first four metrics are automatic metrics, while the last three metrics are human metrics. \textbf{Bold} denotes the best results.}
\label{4}
\end{table*}

\begin{algorithm}[!t]
\caption{GCSD Training Pipeline}
\label{alg:gcsd_training}
\begin{algorithmic}[1]
\State \textbf{Input:} Datasets $D_{\text{real}}, D_{\text{simulated}}$; Base model $M(\theta)$.
\State \textbf{Hyperparams:} SFT weights $\gamma_1, \gamma_2, \gamma_3$; MRPO group size $G$, KL penalty $\beta$.
\State \textbf{Output:} Optimized $\theta_{\text{final}}$.

\Procedure{TrainGCSD}{$D_{\text{real}}, D_{\text{simulated}}, \theta$}
    \State $D_{\text{SFT}} \gets \text{ProcessData}(D_{\text{real}} \cup D_{\text{simulated}})$
    \State $D_{\text{prompt}} \gets \text{ExtractPrompts}(D_{\text{SFT}})$
    \State $\theta_{\text{SFT}} \gets \text{SFT\_Phase}(D_{\text{SFT}}, \theta)$
    \State $\theta_{\text{final}} \gets \text{MRPO\_Phase}(D_{\text{prompt}}, \theta_{\text{SFT}})$
    \State \textbf{return} $\theta_{\text{final}}$
\EndProcedure

\Procedure{SFT\_Phase}{$D_{\text{SFT}}, \theta$}
    \State $\theta_{\text{SFT}} \gets \theta$, $P_{\text{soft, prev}} \gets \mathbf{0}$.
    \For{batch $(U, Y) \in D_{\text{SFT}}$}
    \State $P_{\text{soft, curr}} \gets \text{SoftPromptNet}(\text{UserFeatures}(U))$
    \State $L_{\text{smooth}} \gets \|P_{\text{soft, curr}} - P_{\text{soft, prev}}\|_2^2$
    \State $\text{Logits}, A \gets M_{\theta_{\text{SFT}}}(\text{Concat}(P_{\text{soft, curr}}, \text{Embed}(U)))$
    \State $L_{\text{total}} \gets \gamma_1 \text{CrossEntropyLoss}(\text{Logits}, Y) + \gamma_2 \text{CalculateCSFAL}(A, \text{Keywords}(Y)) + \gamma_3 L_{\text{smooth}}$
    \State Update $\theta_{\text{SFT}}$ via $\nabla L_{\text{total}}$.
    \State $P_{\text{soft, prev}} \gets P_{\text{soft, curr}}$
    \EndFor
    \State \textbf{return} $\theta_{\text{SFT}}$
\EndProcedure

\Procedure{MRPO\_Phase}{$D_{\text{prompt}}, \theta_{\text{SFT}}$}
    \State $\pi_{\theta} \gets \theta_{\text{SFT}}$; $\pi_{\text{ref}} \gets \text{frozen } \theta_{\text{SFT}}$.
    \For{training step}
    \State Sample $q \in D_{\text{prompt}}$.
    \State Generate $\{o_i\}_{i=1}^G \sim \pi_{\theta}(\cdot|q)$.
    \State Compute $R_i \gets \text{CompositeReward}(o_i, \text{reference})$.
    \State Compute advantage estimates $\{\hat{A}_{i,t}\}$.
    \State Compute MRPO objective $J_{\text{MRPO}}(\theta)$.
    \State Update $\theta$ by ascending $\nabla J_{\text{MRPO}}(\theta)$.
    \EndFor
    \State \textbf{return} $\theta$
\EndProcedure

\end{algorithmic}
\end{algorithm}

\noindent
\textbf{3. Smoothness Loss ($\mathcal{L}_{\text{Smoothness}}$):} This term, from the Dynamic Participant Cognitive State Modeling module, ensures the generated soft prompts evolve smoothly over time.

The joint loss ($\mathcal{L}_{\text{SFT}}$) for the supervised fine-tuning phase is a weighted sum of these components:
\begin{equation}
\mathcal{L}_{\text{SFT}} = \gamma_1 * \mathcal{L}_{\text{Gen}} + \gamma_2 * \mathcal{L}_{\text{CSFAL}} + \gamma_3 * \mathcal{L}_{\text{Smoothness}}
\end{equation}
where $\gamma_1, \gamma_2, \text{and } \gamma_3$ are hyperparameters that balance the influence of each loss component.

\subsubsection{Phase 2: Multi-Reward Policy Optimization (MRPO).}


Following SFT, the model's policy is further refined through our multi-reward optimization framework, which is adapted from the GRPO algorithm~\cite{guo2025deepseek}.

For each prompt $q$, we sample a group of $G$ candidate outputs $\{o_1, \dots, o_G\}$ from the current policy $\pi_{\theta}$. The objective is to maximize the following function:

\begin{equation}
\begin{aligned}
J_{\text{MRPO}}(\theta) ={}& \mathbb{E}_{q, \{o_i\}} \left[ \frac{1}{G} \sum_{i=1}^{G} \frac{1}{|o_i|} \sum_{t=1}^{|o_i|} \right. \\
& \left. \left( \frac{\pi_{\theta}(o_{i,t} | q, o_{i,<t})}{\pi_{\theta_{\text{old}}}(o_{i,t} | q, o_{i,<t})} \hat{A}_{i,t} \right. \right. \left. \left. - \beta \cdot \text{KL}_{\text{term}} \right) \right]
\end{aligned}
\end{equation}

\noindent
where $\hat{A}_{i,t}$ is the advantage estimate, $\pi_{\theta_{\text{old}}}$ is the old policy (set to $\pi_{\theta}$ in practice), and $\beta$ is a coefficient controlling the KL penalty. The KL term regularizes the divergence between the policy $\pi_{\theta}$ and a reference model $\pi_{\text{ref}}$ (i.e., the SFT model) to prevent catastrophic forgetting.

\textbf{Multi-Dimensional Reward Mechanism:} The advantage estimate $\hat{A}_{i,t}$ is derived from a rule-based reward model that combines several signals into a single scalar value. To capture established conversational patterns, the model is rewarded for n-gram overlap with reference responses, measured via BLEU-4. To encourage the learning of latent semantics and principles, we evaluate semantic similarity using BERTScore. Response diversity, which is crucial for engagement, is promoted by rewarding unique tokens, as quantified by Distinct-2 metrics. Finally, a binary reward enforces structural correctness, such as the proper use of \texttt{[Assistant]} tokens. This composite reward is the core mechanism for enforcing the 'principle-driven' aspect of our policy, driving the MRPO process to optimize the model’s outputs for cognitive stimulation relevance and quality.

\section{Experiments}

\subsection{Implementation Details}

The base model of the GCSD is Qwen-2.5-3b\footnote{\url{https://huggingface.co/Qwen/Qwen2.5-3B-Instruct}}. All models are implemented with PyTorch~\cite{paszke2019pytorch} on one NVIDIA A100-80G GPU. Training is orchestrated using the AdamW optimizer~\cite{loshchilov2017decoupled} with decoupled weight decay ($\lambda=0.01$). We employ a \textit{Linear Warmup with Cosine Annealing Learning Rate Schedule}~\cite{waswani2017attention}, where the learning rate linearly increases from $10^{-7}$ to a peak of $5 \times 10^{-5}$ over the initial 5\% of training steps, subsequently annealing down to $10^{-7}$ following a cosine schedule. A micro-batch size of 1 is utilized, accumulated to an effective batch size of 16 through gradient accumulation steps. Furthermore, \textit{mixed-precision training (FP16)} is employed to accelerate computation and reduce memory footprint. 
During inference, a temperature of 0.6 and a top-$p$ value of 0.95 are set for stochastic decoding, balancing generation diversity with coherence, consistent with best practices in large language model inference~\cite{liu2024deepseek, guo2025deepseek, shao2024deepseekmath}. For evaluation, the ROUGE-chinese package (version 1.0.3) is used to compute the ROUGE metric, the NLTK package (version 3.9.1) is used to compute the Bleu metric, and the bert-score package (version 0.3.10) is used to compute BERTScore.

\subsection{Evaluation}
\subsubsection{Automatic Evaluation.}
In the generation of cognitive stimulation dialogues, we utilize the ROUGE-L metric~\cite{lin2004ROUGE, ganesan2018ROUGE} and BLEU-4 metric~\cite{papineni2002bleu} to evaluate the quality of the text. In order to better evaluate semantic similarity, we also implement BERTScore~\cite{zhang2019bertscore}. BERTScore assists in measuring the similarity between the embeddings of the generated sentence and the reference sentence. In addition, we employ Distinct~\cite{li2015diversity}, which quantify the proportion of uni-gram, respectively, across all generated outputs, thereby providing an indicator of diversity.

\subsubsection{Human Evaluation.}

To qualitatively examine model performance, human evaluations are conducted. Dialogues from the GCSD and the baselines are sampled. Six elders and their relatives participate in evaluating the responses generated by different models. Following~\cite{lietal2020empdg, liuetal2021towards, li2022knowledge, jiang2023cognitive}, all models undergo evaluation in terms of \textbf{Relevance}, \textbf{Empathy}, \textbf{Fluency}. \textbf{Relevance} assesses whether the generated responses align with the dialogue context. \textbf{Empathy} determines if the LISTENER understands the SPEAKER's feelings. \textbf{Fluency} assesses the grammatical correctness and readability of the SPEAKER's responses. Each metric is rated on a five-point scale, where 1, 3, and 5 represent unacceptable, moderate, and excellent performance, respectively.

\subsection{Baselines}

For a comprehensive comparison, we evaluate our GCSD system against a suite of powerful, mainstream large language models. These include our base model, \textbf{Qwen-2.5-3b}~\cite{yang2025qwen2}, as well as other prominent models such as Baidu's \textbf{ERNIE}~\cite{zhang2019ernie}, ByteDance's \textbf{Doubao-Pro}\footnote{https://www.doubao.com/chat/}, \textbf{DeepSeek-671B}~\cite{liu2024deepseek}, Meta's \textbf{LLaMA-3.1-405B}~\cite{dubey2024llama}, Zhipu AI's \textbf{GLM-4-Plus}~\cite{glm2024chatglm}, Alibaba's \textbf{Qwen-Max}~\cite{bai2023qwen}, and OpenAI's \textbf{ChatGPT}\footnote{https://chatgpt.com/} and its flagship model, \textbf{GPT-4o}~\cite{hurst2024gpt}.

To validate the contribution of each component and our data strategy, we conduct an ablation study. \textbf{w/o CT} bypasses the initial training on simulated data, fine-tuning the base model exclusively on the real dataset. \textbf{w/o DPSM} ablates the dynamic participant state modeling module. \textbf{w/o CSFAL} excludes the cognitive stimulation-focused attention loss. 


\section{Results}

\subsection{Experimental Results}

Our GCSD model achieves state-of-the-art performance on the real-world dataset, consistently outperforming strong baseline models (Table~\ref{4}). The core insight from the results is that GCSD's primary advantage stems not merely from generating semantically relevant content, but from its mastery of the specific, principle-guided interaction structures inherent to cognitive stimulation therapy. This is quantitatively demonstrated by its \textbf{BLEU-4} score of \textbf{27.93}, a \textbf{14.7\%} improvement over the strongest 5-shot baseline, providing clear evidence that it effectively learns cognitive stimulation phrasing and multi-party turn-taking patterns. This structural advantage is particularly significant when contrasted with semantic similarity, where the gap between GCSD (\textbf{BERTScore}: \textbf{80.12}) and the top baseline (79.98) is narrow. This structural command is further validated by human evaluations, where GCSD attains the highest score in Relevance (4.15), a critical metric for multi-party dialogue that indicates the model successfully overcomes challenges like speaker confusion. Ultimately, human A/B testing confirms its practical superiority, showing a decisive win rate against established models like ERNIE (75\%) and a net positive preference over the powerful GPT-4o (50\% win vs. 39\% loss), validating its suitability for this specialized domain (Table~\ref{TableH}).

The ablation study confirms that GCSD's effectiveness stems from the synergistic integration of its components, rather than any single module. 
Bypassing the simulated data pre-training (w/o CT) leads to a noticeable performance drop (\textbf{BLEU-4} from \textbf{27.93} to \textbf{26.51}), highlighting the value of the dual-data strategy in teaching the model the foundational framework of CST. The performance degradation observed when removing the dynamic participant state modeling (w/o DPSM) and the cognitive stimulation-focused attention loss (w/o CSFAL) also confirms their respective contributions to personalized interaction and therapeutic reasoning. Therefore, GCSD's success is not attributable to a single ``silver bullet'' but to the methodical integration of all four complementary modules.

\begin{table}[h!]
\centering
\begin{tabular}{l|c|c|c}
\toprule
\textbf{Models} & \textbf{Win} & \textbf{Loss} & \textbf{Tie}\\
\midrule
GCSD-3b vs ERNIE & 75\% & 10\% & 15\%\\
GCSD-3b vs GLM-4-Plus & 73\% & 12\% & 15\%\\
GCSD-3b vs Doubao-Pro & 60\% & 23\% & 17\% \\
GCSD-3b vs ChatGPT & 58\% & 27\% & 15\% \\
GCSD-3b vs GPT-4o & 50\% & 39\% & 11\% \\
GCSD-3b vs LLaMA-3.1-405B & 58\% & 34\% & 8\% \\
GCSD-3b vs DeepSeek-671B & 43\% & 40\% & 17\% \\
GCSD-3b vs Qwen-Max & 55\% & 29\% & 16\% \\
\bottomrule
\end{tabular}
\caption{Result of human A/B test.}
\label{TableH}
\end{table}

\subsection{Ablation Study}


The results of the ablation study (Table~\ref{4}) clearly demonstrate that GCSD's superior performance stems from the synergistic effect of its integrated components, rather than the contribution of any single module. 
In addition, the value of our dual-data strategy is confirmed, as removing the simulated data pre-training (w/o CT) resulted in a competitive but lower BLEU-4 score of 26.51, highlighting this phase's role as a crucial ``primer'' for the model.

The study also validated the contributions of the more granular architectural components. Removing the Dynamic Participant Cognitive State Modeling (w/o DPSM) led to a noticeable performance drop (BLEU-4 to 23.15), confirming its importance in achieving personalized and adaptive interactions. Similarly, excluding the cognitive stimulation-focused Attention Loss (w/o CSFAL) also resulted in a lower BLEU-4 score of 24.98, showing its effectiveness in instilling the necessary therapeutic reasoning. In summary, GCSD's leading performance is not due to a single ``silver bullet'' but is the result of the methodical integration and synergy of all four complementary components.

\section{Conclusion and Outlook}



This paper proposes a principle-driven adaptive policy for multi-party cognitive stimulation dialogue, namely GCSD. By leveraging a large-scale real and simulated dataset and four core model modules, GCSD addresses the challenges of multi-party dialogue management, personalized cognitive state modeling, and alignment with cognitive stimulation principles. Experimental results show that GCSD achieves strong performance on both automatic and human evaluation in cognitive stimulation dialogue.

Looking ahead, our future work will proceed in three directions. First, we aim to conduct long-term clinical trials to validate the therapeutic efficacy of GCSD in real-world care settings, bridging the gap between computational metrics and clinical outcomes. Second, we plan to extend the current text-based framework to a multimodal system that incorporates visual and acoustic cues, enabling more empathetic and responsive interactions for elders with varying degrees of impairment. Finally, we will investigate safety mechanisms to prevent hallucinations in medical contexts, ensuring that the system remains a reliable and ethical companion for cognitive health support.

\section{Acknowledgments}
We want to thank our anonymous AC, SPC and reviewers for their feedback. We acknowledge support from The University of Hong Kong and Guangdong Laboratory of Artificial Intelligence and Digital Economy (SZ). This work was supported by Hong Kong Innovation and Technology Commission’s Innovation and Technology Fund (Award No. ITS/269/22FP).

\bibliography{aaai2026}

\end{document}